# Rolling Shutter Camera Relative Pose: Generalized Epipolar Geometry

Yuchao Dai[1], Hongdong Li[1,2] and Laurent Kneip[1,2]
[1] Research School of Engineering, Australian National University
[2] ARC Centre of Excellence for Robotic Vision (ACRV)

## Abstract

*The vast majority of modern consumer-grade cameras employ a rolling shutter mechanism. In dynamic geometric computer vision applications such as visual SLAM, the so-called rolling shutter effect therefore needs to be properly taken into account. A dedicated relative pose solver appears to be the first problem to solve, as it is of eminent importance to bootstrap any derivation of multi-view geometry. However, despite its significance, it has received inadequate attention to date.*

*This paper presents a detailed investigation of the geometry of the rolling shutter relative pose problem. We introduce the rolling shutter essential matrix, and establish its link to existing models such as the push-broom cameras, summarized in a clean hierarchy of multi-perspective cameras. The generalization of well-established concepts from epipolar geometry is completed by a definition of the Sampson distance in the rolling shutter case. The work is concluded with a careful investigation of the introduced epipolar geometry for rolling shutter cameras on several dedicated benchmarks.*

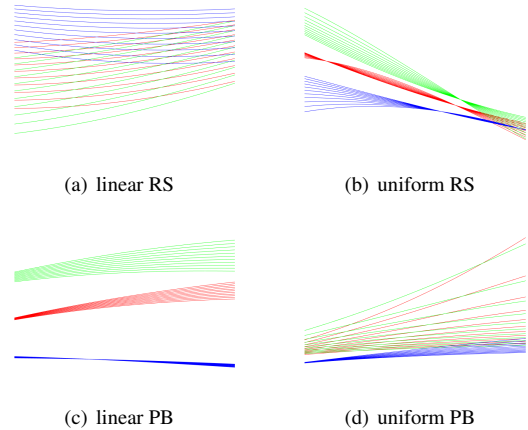

Figure 1. Example epipolar curves for the camera models discussed in this paper. Groups of epipolar curves of identical color originate from points on the same row in another image, while both images are under motion. For linear rolling shutter (a) and linear push broom cameras (c), the epipolar curves are conic. The epipolar curves for uniform rolling shutter (b) and uniform push broom cameras (d) are cubic.

## 1. Introduction

Rolling-Shutter (RS) CMOS cameras are getting more and more popularly used in real-world computer vision applications due to their low cost and simplicity in design. To use these cameras in 3D geometric computer vision tasks (such as 3D reconstruction, object pose, visual SLAM), the rolling shutter effect (*e.g.* wobbling) must be carefully accounted for. Simply ignoring this effect and relying on a global-shutter method may lead to erroneous, undesirable and distorted results as reported in previous work (*e.g.* [11, 13, 3]).

Recently, many classic 3D vision algorithms have been adapted to the rolling shutter case (*e.g.* absolute Pose [15] [3] [22], Bundle Adjustment [9], and stereo rectification [21]). Quite surprisingly, no previous attempt has been reported on solving the **relative pose** problem with a Rolling Shutter (RS) camera.

The complexity of this problem stems from the fact that a rolling shutter camera does not satisfy the pinhole projection model, hence the conventional epipolar geometry defined by the standard $3 \times 3$ essential matrix (in the form of $\mathbf{x}'^T \mathbf{E} \mathbf{x} = 0$) is no longer applicable. This is mainly because of the time-varying scaneline-by-scanline image capturing nature of an RS camera, rendering the imaging process a non-central one.

In this paper we show that similar epipolar relationships do exist between two rolling-shutter images. Specifically, in contrast to the conventional $3 \times 3$ essential matrix for the pinhole camera, we derive a $7 \times 7$ generalized essential matrix for a *uniform rolling-shutter* camera, and a $5 \times 5$ generalized essential matrix for a *linear rolling-shutter* camera. Another result is that, under the rolling-shutter epipolar geometry, the "epipolar lines" are no longer straight lines, but become higher-order "epipolar curves" (*c.f.* Fig. 1).

Armed with these novel generalized rolling-shutter es-sential matrices, we can easily develop efficient numerical algorithms to solve the rolling shutter relative pose problem. Similar to the 8-point linear algorithm in the perspective case, we derive a 20-point linear algorithm for linear RS cameras, and a 44-point linear algorithm for uniform RS cameras. We also develop non-linear solvers for both cases (by minimizing the geometrically meaningful Sampson error). Our non-linear solvers work for the minimum number of feature points, hence are relevant for RANSAC.

Table 1. A hierarchy of generalized essential matrices for different types of rolling-shutter and push-broom cameras.

| Camera Model | Essential Matrix | Monomials | Degree-of-freedom | Linear Algorithm | Non-linear Algorithm | Motion Parameters |
|---|---|---|---|---|---|---|
| Perspective camera | $\begin{bmatrix} f_{11} & f_{12} & f_{13} \\ f_{21} & f_{22} & f_{23} \\ f_{31} & f_{32} & f_{33} \end{bmatrix}$ | $(u_i, v_i, 1)$ | $3^2 = 9$ | 8-point | 5-point | $\mathbf{R}, \mathbf{t}$ |
| Linear push broom | $\begin{bmatrix} 0 & 0 & f_{13} & f_{14} \\ 0 & 0 & f_{23} & f_{24} \\ f_{31} & f_{32} & f_{33} & f_{34} \\ f_{41} & f_{42} & f_{43} & f_{44} \end{bmatrix}$ | $(u_i v_i, u_i, v_i, 1)$ | $12 = 4^2 - 2^2$ | 11-point | 11-point | $\mathbf{R}, \mathbf{t}, \mathbf{d}_1, \mathbf{d}_2$ |
| Linear rolling shutter | $\begin{bmatrix} 0 & 0 & f_{13} & f_{14} & f_{15} \\ 0 & 0 & f_{23} & f_{24} & f_{25} \\ f_{31} & f_{32} & f_{33} & f_{34} & f_{35} \\ f_{41} & f_{42} & f_{43} & f_{44} & f_{45} \\ f_{51} & f_{52} & f_{53} & f_{54} & f_{55} \end{bmatrix}$ | $(u_i^2, u_i v_i, u_i, v_i, 1)$ | $21 = 5^2 - 2^2$ | 20-point | 11-point | $\mathbf{R}, \mathbf{t}, \mathbf{d}_1, \mathbf{d}_2$ |
| Uniform push broom | $\begin{bmatrix} 0 & 0 & f_{13} & f_{14} & f_{15} & f_{16} \\ 0 & 0 & f_{23} & f_{24} & f_{25} & f_{26} \\ f_{31} & f_{32} & f_{33} & f_{34} & f_{35} & f_{36} \\ f_{41} & f_{42} & f_{43} & f_{44} & f_{45} & f_{46} \\ f_{51} & f_{52} & f_{53} & f_{54} & f_{55} & f_{56} \\ f_{61} & f_{62} & f_{63} & f_{64} & f_{65} & f_{66} \end{bmatrix}$ | $(u_i^2 v_i, u_i^2, u_i v_i, u_i, v_i, 1)$ | $32 = 6^2 - 2^2$ | 31-point | 17-point | $\mathbf{R}, \mathbf{t}, \mathbf{w}_1, \mathbf{w}_2, \mathbf{d}_1, \mathbf{d}_2$ |
| Uniform rolling shutter | $\begin{bmatrix} 0 & 0 & f_{13} & f_{14} & f_{15} & f_{16} & f_{17} \\ 0 & 0 & f_{23} & f_{24} & f_{25} & f_{26} & f_{27} \\ f_{31} & f_{32} & f_{33} & f_{34} & f_{35} & f_{36} & f_{37} \\ f_{41} & f_{42} & f_{43} & f_{44} & f_{45} & f_{46} & f_{47} \\ f_{51} & f_{52} & f_{53} & f_{54} & f_{55} & f_{56} & f_{57} \\ f_{61} & f_{62} & f_{63} & f_{64} & f_{65} & f_{66} & f_{67} \\ f_{71} & f_{72} & f_{73} & f_{74} & f_{75} & f_{76} & f_{77} \end{bmatrix}$ | $(u_i^3, u_i^2 v_i, u_i^2, u_i v_i, u_i, v_i, 1)$ | $45 = 7^2 - 2^2$ | 44-point | 17-point | $\mathbf{R}, \mathbf{t}, \mathbf{w}_1, \mathbf{w}_2, \mathbf{d}_1, \mathbf{d}_2$ |

Experiments on both synthetic RS datasets and real RS images have validated the proposed theory and algorithms. To the best of our knowledge, this is the first work that provides a unified framework and practical solutions to the rolling shutter relative pose problem. Our $5 \times 5$ and $7 \times 7$ RS essential matrices are original; they were not reported before in computer vision literature. Inspired by this success, we further discover that there also exist practically meaningful $4 \times 4$ and $6 \times 6$ generalized essential matrices, corresponding to linear, and uniform push-broom cameras, respectively. Together, this paper provides a unified framework for solving the relative pose problems with rolling-shutter or push-broom cameras under different yet practically relevant conditions. It also provides new geometric insights into the connection between different types of novel camera geometries.

*Table-1* gives a brief summary of the new results discovered in this paper. Details will be explained in Section-4.

### 1.1. Related work

The present work discusses a fundamental geometric problem in the context of rolling shutter cameras. The most notable, early related work is by Geyer et al. [16], which proposes a projection model for rolling shutter cameras based on a constant velocity motion model. This fundamental idea of a compact, local expression of camera dynamics has regained interest through Ait-Aider et al. [1], who solved the absolute pose problem through iterative minimization, and for the first time described the higher density of the temporal sampling of a rolling shutter mechanism as an advantage rather than a disadvantage. Albl et al.[3] proposed a two-step procedure in which the pose is first initialized using a global shutter model, and then refined based on a rolling shutter model and a small-rotation approximation. Saurer et al. [22] solved the problem in a single shot, however under the simplifying assumption that the rotational velocity of the camera is zero. Sunghoon et al. [11] also employed a linear model, however with the final goal of dense depth estimation from stereo. Grundmann et al. proposed a method to automatic rectify rolling shutter distortion from feature correspondences only [5]. To date, a single-shot, closed-form solution to compute the relative pose for a rolling shutter camera remains an open problem, thus underlining the difficulty of the geometry even in the first-order case.

Rolling shutter cameras can be regarded as general multi-perspective cameras, and are thus closely related to several other camera models. For instance, Gupta and Hartley [6] introduced the linear push-broom model where—similar to rolling shutter cameras—the vertical image coordinate becomes correlated to the time at which the corresponding row is sampled. This notably leads to a quadratic essential polynomial and a related, higher-order essential matrix. We establish the close link to this model and contribute to the classification in [27] by presenting a novel hierarchy of higher order generalized essential matrices.

Moving towards iterative non-linear refinement methods permits a more general inclusion of higher-order motion models. Hedborg et al. [9, 10] introduced a bundle adjustment framework for rolling shutter cameras by relying on the SLERP model for interpolating rotations. Magarand et al. [15] introduced an approach for global optimization of pose and dynamics from a single rolling shutter image.

Oth et al. [17] proposed to use more general temporal basis functions for parameterizing the trajectory of the camera.

Solutions to the rolling shutter problem have also been explored for further types of sensors. For instance, Ait-Aider and Berry [2] and Saurer et al. [21] had anaylzed the problem in the context of stereo cameras. Recently, Kerl et al. [13] have started to apply continuous time parametrizations to RGB-D cameras. Ponce studied general concept of various types of cameras including line pencil camera akin to general rolling shutter [20].

The relative pose problem is of eminent importance in structure-from-motion, as it allows to bootstrap the computation in the absence of any information about the structure. To the best of our knowledge, the present work is the first to address it in the context of a rolling shutter camera.

## 2. Rolling-Shutter Camera Models

A critical difference between a rolling shutter camera and a pinhole camera is that the former does no longer possess a single center-of-projection in the general case. Instead, in a rolling-shutter image, each of its scanlines generally has a different effective projection center (**temporal-dynamic**) as well as a different local frame and orientation.

In an attempt to present this matter from a mathematical perspective, let us start with re-examining the model of a (global shutter) pinhole camera, which can be entirely described by a central projection matrix: $\mathbf{P} = \mathbf{K}[\mathbf{R}, \mathbf{t}]$.

When an RS camera is in motion during image acquisition, all its scanlines are sequentially exposed at different time steps; hence each scanline possesses a different local frame. Mathematically, we need to assign a unique projection matrix to every scanline in an RS image. For example, for the $u_i$-th scanline, we have

$$\mathbf{P}_{u_i} = \mathbf{K}[\mathbf{R}_{u_i}, \mathbf{t}_{u_i}]. \quad (1)$$

### 2.1. Rolling-Shutter Model Classification

**General Rolling Shutter Camera.** It is common to assume that the motion of a rolling-shutter camera during image acquisition is smooth. Otherwise an arbitrarily non-smoothly moving RS camera would create meaningless images suffering from arbitrary fragmentations.

Therefore, a smoothly moving RS camera is considered as the most general form of rolling-shutter models. It is easy to see that for a general RS image its scanlines' local pose matrices $\mathbf{P_0}, \mathbf{P_1}, \mathbf{P_2}, \cdots, \mathbf{P_{N-1}}$ will trace out a smooth trajectory in the $\mathbf{SE}(3)$ space. B-splines have been used to model this trajectory in the RS context [24, 18, 13].

To ease the derivation, we assume the RS camera is intrinsically calibrated (*i.e.*, $\mathbf{K}$ is assumed to be known, and can be omitted thereafter). However, note that many of the results presented in this paper remain extendable to the uncalibrated case as well (by transitioning from the essential matrix to the corresponding fundamental matrix). Also note that the task of intrinsic calibration can be easily done, *e.g.* by applying any standard camera calibration procedure to still imagery taken by an RS camera.

**Linear Rolling-Shutter Camera.** The motion of the camera is a pure translation by a constant linear velocity. The orientations of local scanline frames are constant. In this case, the projection centers of the scanlines lie on a straight line in 3D space. Supposing that constant velocity induces a translation shift of $\mathbf{d}$ per image row (expressed in normalized coordinates), we can write down the $u_i$-th projection matrix as

$$\mathbf{P}_{u_i} = [\mathbf{R}_0, \mathbf{t}_0 + u_i \mathbf{d}]. \quad (2)$$

We use the top-most scanline's local frame $[\mathbf{R}_0, \mathbf{t}_0]$ as the reference frame of the RS image.

**Uniform Rolling-Shutter Camera.** The uniform rolling-shutter camera is another popular RS model, which is more general than the linear RS model. The camera is performing a uniform rotation at a constant angular velocity, in addition to a uniform linear translation at constant linear velocity. All the centers of projection form a helix spiral trajectory.

We use $\mathbf{d} \in \mathbb{R}^3$ to denote the constant linear velocity and $\mathbf{w} \in \mathbb{R}^3$ for the constant angular velocity (expressing angular displacement per row). Let $\mathbf{w}$ be parametrized in the angle-axis representation (*i.e.* $\mathbf{w} = \omega[\mathbf{n}]$). The $u_i$-th scanline's local projection matrix is $\mathbf{P}_{u_i} = [\mathbf{R}_{u_i}, \mathbf{t}_{u_i}]$, where

$$\begin{aligned} \mathbf{R}_{u_i} &= (\mathbf{I} + \sin(u_i\omega)[\mathbf{n}]_\times + (1 - \cos(u_i\omega))[\mathbf{n}]_\times^2)\mathbf{R}_0, \\ \mathbf{t}_{u_i} &= \mathbf{t}_0 + u_i \mathbf{d}. \end{aligned} \quad (3)$$

One may further assume that the inter-scanline rotation during image acquisition is very small. This is a reasonable assumption, as the acquisition time for a single image is very short, often in the order of 10s milliseconds, and the motion of an RS camera is typically small. Under the small-rotation approximation, we have

$$\begin{aligned} \mathbf{R}_{u_i} &= (\mathbf{I} + u_i\omega[\mathbf{n}]_\times)\mathbf{R}_0, \\ \mathbf{t}_{u_i} &= \mathbf{t}_0 + u_i \mathbf{d}. \end{aligned} \quad (4)$$

## 3. The Rolling Shutter Relative Pose Problem

The RS Relative Pose problem consists of finding the relative camera displacement between two RS views, given image feature correspondences.

It is well known that for the perspective case the epipolar geometry plays a central role in relative pose estimation, translated into a simple 3-by-3 matrix called the essential (or fundamental) matrix. Specifically, given a set of correspondences between two views, $\mathbf{x}_i = [u_i, v_i, 1]^T \leftrightarrow \mathbf{x}'_i =$

$[u'_i, v'_i, 1]^T$, we have the standard essential matrix constraint: $\mathbf{x}'^T_i \mathbf{E} \mathbf{x}_i = 0$. From a sufficient number of correspondences one can solve for $\mathbf{E}$. Once $\mathbf{E}$ is obtained, decomposing $\mathbf{E}$ according to $\mathbf{E} = [\mathbf{t}]_\times \mathbf{R}$ leads to the relative pose (*i.e.* $\mathbf{R}$ and $\mathbf{t}$).

For a rolling-shutter camera, unfortunately, such a global 3-by-3 essential matrix does not exist. This is primarily because an RS camera is not a central projection camera; every scanline has its own distinct local pose. As a result, every pair of feature correspondences may give rise to a different "essential matrix". Formally, for $\mathbf{x}_i \leftrightarrow \mathbf{x}'_i$, we have

$$\mathbf{x}'^T_i \mathbf{E}_{u_i, u'_i} \mathbf{x}_i = 0. \quad (5)$$

Note that $\mathbf{E}$ is dependent of the scanlines $u_i$ and $u'_i$. In other words, there does not exist a single global $3 \times 3$ essential matrix for a pair of RS images.

Figure-2 shows that despite the fact that different scanlines possess different centers of projection, for a pair of feature correspondences the co-planarity relationship still holds, because the two feature points in image planes correspond to the same 3D point in space. As such, the concept of two-view epipolar relationship should still exist. Our next task is to derive such a generalized epipolar relation.

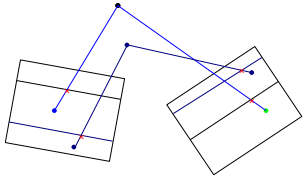

Figure 2. This figure shows that different scanlines in a RS image have different effective optical centers. For any pair of feature correspondences (indicated by red 'x's in the picture), a co-planarity relationship however still holds.

Given two scanlines $u_i, u_j$ and the corresponding camera poses $\mathbf{P}_{u_i} = [\mathbf{R}_{u_i}, \mathbf{t}_{u_i}]$ and $\mathbf{P}_{u_j} = [\mathbf{R}_{u_j}, \mathbf{t}_{u_j}]$, we have

$$\mathbf{E}_{u_i u_j} = [\mathbf{t}_{u_j} - \mathbf{R}_{u_j} \mathbf{R}^T_{u_i} \mathbf{t}_{u_i}]_\times \mathbf{R}_{u_j} \mathbf{R}^T_{u_i}. \quad (6)$$

**Rolling Shutter Relative Pose.** Note, given a pair of feature correspondences $\mathbf{x}_i \leftrightarrow \mathbf{x}'_i$, one can establish the following RS epipolar equation: $\mathbf{x}'^T_i \mathbf{E}_{u_i u'_i} \mathbf{x}_i = 0$. Given sufficient pairs of correspondences; each pair contributes to one equation over the unknown parameters; our goal is to solve for the relative pose between the two RS images.

We set the first camera's pose at $[\mathbf{I}, \mathbf{0}]$, and the second camera at $[\mathbf{R}, \mathbf{t}]$. We denote the two cameras' inter-scanline rotational (angular) velocities as $\mathbf{w}_1$, and $\mathbf{w}_2$, and their linear translational velocities as $\mathbf{d}_1$ and $\mathbf{d}_2$. Taking a uniform RS camera as an example, the task of rolling shutter relative pose is to find the unknowns $\{\mathbf{R}, \mathbf{t}, \mathbf{w}_1, \mathbf{w}_2, \mathbf{d}_1, \mathbf{d}_2\}$.

In total there are $2 \times 12 - 6 - 1 = 17$ non-trivial variables (excluding the gauge freedom of the first camera, and a global scale). Collecting at least 17 equations in general configuration, it is possible to solve this system of (generally nonlinear) equations over the 17 unknown parameters. In this paper, we will show how to derive *linear N-point algorithms* for rolling shutter cameras, as an analogy to the linear 8-point algorithm for the case of a pinhole camera.

## 4. Rolling-Shutter Essential Matrices

In this section, we will generalize the conventional $3 \times 3$ essential matrix for perspective cameras to $4 \times 4, 5 \times 5, 6 \times 6$, and $7 \times 7$ matrices for different types of Rolling-Shutter (RS) and Push-Broom (PB) cameras. The reason for including push-broom cameras will be made clear soon.

### 4.1. A $5 \times 5$ essential matrix for linear RS cameras

For a linear rolling shutter camera, since the inter-scanline motion is a pure translation, there are four parameter vectors to be estimated, namely $\{\mathbf{R}, \mathbf{t}, \mathbf{d}_1, \mathbf{d}_2\}$. The total degree of freedom of the unknowns is $3+3+3+3-1 = 11$. (the last '-1' accounts for a global scale).

The epipolarity defined between the $u_i$-th scanline of the first RS frame and the $u'_i$-th scanline of the second RS frame is represented as $\mathbf{E}_{u_i u'_i} = [\mathbf{t}_{u_i u'_i}]_\times \mathbf{R}_{u_i u'_i}$, where the translation $\mathbf{t}_{u_i u'_i} = \mathbf{t} + u'_i \mathbf{d}_2 - u_i \mathbf{R} \mathbf{d}_1$. This translates into

$$\begin{bmatrix} u'_i \\ v'_i \\ 1 \end{bmatrix}^T [\mathbf{t} + u'_i \mathbf{d}_2 - u_i \mathbf{R} \mathbf{d}_1]_\times \mathbf{R} \begin{bmatrix} u_i \\ v_i \\ 1 \end{bmatrix} = 0. \quad (7)$$

Expanding this scanline epipolar equation, one can obtain the following $5 \times 5$ matrix form:

$$\begin{bmatrix} u'^2_i \\ u'_i v'_i \\ u'_i \\ v'_i \\ 1 \end{bmatrix}^T \begin{bmatrix} 0 & 0 & f_{13} & f_{14} & f_{15} \\ 0 & 0 & f_{23} & f_{24} & f_{25} \\ f_{31} & f_{32} & f_{33} & f_{34} & f_{35} \\ f_{41} & f_{42} & f_{43} & f_{44} & f_{45} \\ f_{51} & f_{52} & f_{53} & f_{54} & f_{55} \end{bmatrix} \begin{bmatrix} u^2_i \\ u_i v_i \\ u_i \\ v_i \\ 1 \end{bmatrix} = 0, \quad (8)$$

where the entries of the $5 \times 5$ matrix $\mathbf{F} = [f_{i,j}]$ are functions of the 11 unknown parameters $\{\mathbf{R}, \mathbf{t}, \mathbf{d}_1, \mathbf{d}_2\}$. In total, there are 21 homogeneous variables, thus a linear 20-point solver must exist to solve for this hyperbolic essential matrix.

*Proof.* By redefining $\mathbf{d}_1 \leftarrow \mathbf{R} \mathbf{d}_1$, we easily obtain

$$\mathbf{E}_{u_i u'_i} = \left([\mathbf{t}]_\times + u'_i [\mathbf{d}_2]_\times - u_i [\mathbf{d}_1]_\times\right) \mathbf{R}. \quad (9)$$

Denoting $\mathbf{E}_0 = [\mathbf{t}]_\times \mathbf{R}, \mathbf{E}_1 = [\mathbf{d}_1]_\times \mathbf{R}$ and $\mathbf{E}_2 = [\mathbf{d}_2]_\times \mathbf{R}$, we have:

$$[u'_i, v'_i, 1](\mathbf{E}_0 + u'_i \mathbf{E}_2 - u_i \mathbf{E}_1)[u_i, v_i, 1]^T = 0. \quad (10)$$

The $5 \times 5$ matrix $\mathbf{F}$ is defined in the following way

$$\mathbf{F} = \begin{bmatrix} 0 & 0 & E_{1,11} & E_{1,21} & E_{1,31} \\ 0 & 0 & E_{1,12} & E_{1,22} & E_{1,32} \\ E_{2,11} & E_{2,21} & a & b & c \\ E_{2,12} & E_{2,22} & E_{0,12}+E_{2,32} & E_{0,22} & E_{0,32} \\ E_{2,13} & E_{2,23} & E_{0,13}+E_{2,33} & E_{0,23} & E_{0,33} \end{bmatrix}, \quad (11)$$

where $a = E_{0,11} + E_{1,13} + E_{2,31}, b = E_{0,21} + E_{1,23}, c = E_{0,31} + E_{1,33}$. Finally, it is easy to verify the equation

$$[u_i'^2, u_i'v_i', u_i', v_i', 1]\mathbf{F}[u_i^2, u_iv_i, u_i, v_i, 1]^T = 0.$$

□

**Hyperbolic epipolar curves.** Note that the "epipolar lines" for a linear RS camera are hyperbolic curves. It is easy to verify that the generalized essential matrix for linear rolling shutter camera is full rank and the epipole lies in infinity.

**Difference with axial camera** An axial camera is a particular case of non-central cameras where all the back-projection rays intersect a line in 3D (the axis) [26]. The linear rolling shutter camera give rise to an axis where every back-projection ray intersects (center of projection). However, the temporal-dynamic nature of linear rolling shutter camera distinguishes itself from the axial camera [26], where the internal displacement (linear velocity) is unknown and to estimate. Even though our linear RS essential matrix shares the same size as axial camera essential matrix [25], the detailed structure is different.

### 4.2. A $7 \times 7$ essential matrix for uniform RS cameras

Consider a uniform RS camera undergoing a rotation at constant angular velocity $\mathbf{w}$ and a translation at constant linear velocity $\mathbf{d}$. We assume the angular velocity is very small. By using the small-rotation approximation, we have the $u_i$-th scanline's local pose as

$$\mathbf{P}_{u_i} = [(\mathbf{I} + u_i[\mathbf{w}]_\times)\mathbf{R}_0, \ \mathbf{t}_0 + u_i\mathbf{d}]. \quad (12)$$

Given a pair of two corresponding uniform RS camera frames, we then have

$$[u_i', v_i', 1][\mathbf{t} + u_i'\mathbf{d}_2 - u_i\mathbf{R}_{u_i u_i'}\mathbf{d}_1]_\times \mathbf{R}_{u_i u_i'}[u_i, v_i, 1]^T = 0, \quad (13)$$

Expanding this equation with the aid of the small rotation approximation results in

$$\mathbf{R}_{u_i, u_i'} = (\mathbf{I} + u_i'[\mathbf{w}_2]_\times)\mathbf{R}_0(\mathbf{I} - u_i[\mathbf{w}_1]_\times), \quad (14)$$

and we finally obtain:

$$\left[u_i'^3, u_i'^2v_i', u_i'^2, u_i'v_i', u_i', v_i', 1\right] \mathbf{F} \left[u_i^3, u_i^2v_i, u_i^2, u_iv_i, u_i, v_i, 1\right]^T = 0, \quad (15)$$

where

$$\mathbf{F} = \begin{bmatrix} 0 & 0 & f_{13} & f_{14} & f_{15} & f_{16} & f_{17} \\ 0 & 0 & f_{23} & f_{24} & f_{25} & f_{26} & f_{27} \\ f_{31} & f_{32} & f_{33} & f_{34} & f_{35} & f_{36} & f_{37} \\ f_{41} & f_{42} & f_{43} & f_{44} & f_{45} & f_{46} & f_{47} \\ f_{51} & f_{52} & f_{53} & f_{54} & f_{55} & f_{56} & f_{57} \\ f_{61} & f_{62} & f_{63} & f_{64} & f_{65} & f_{66} & f_{67} \\ f_{71} & f_{72} & f_{73} & f_{74} & f_{75} & f_{76} & f_{77} \end{bmatrix}.$$

This gives a $7 \times 7$ RS essential matrix $\mathbf{F}$, whose elements are functions of the 18 unknowns (*i.e.* $\{\mathbf{R}, \mathbf{t}, \mathbf{w}_1, \mathbf{w}_2, \mathbf{d}_1, \mathbf{d}_2\}$). Also note the induced epipolar curves are *cubic*.

In total there are 45 homogeneous variables, thus a 44-point linear algorithm exists to solve for this hyperbolic essential matrix. The generalized essential matrix for uniform rolling shutter camera is full rank and the epipole lies in infinity.

**Reducing to a linear RS camera.** Under the above formulation, if $\mathbf{w}_1 = \mathbf{w}_2 = \mathbf{0}$, the equation will reduce to Eq.-(10), *i.e.*, the linear rolling shutter case.

### 4.3. A $4 \times 4$ essential matrix for linear PB cameras

Researchers have previously noticed the similarity between a spacetime sweeping camera (such as RS) and a push-broom camera [16, 23]. Here, we further illustrate this similarity, via our high-order essential matrix.

Specifically, the above $5 \times 5$ and $7 \times 7$ RS essential matrices have inspired us to explore further. Do $4 \times 4$ or $6 \times 6$ generalized essential matrices also exist? Following a similar approach, we quickly find out that: these two generalized essential matrices do exist and they each corresponds to a special type of push-broom camera.

For linear push-broom (PB) cameras (as defined in [6]), there exists a $4 \times 4$ essential matrix:

$$\mathbf{F} = \begin{bmatrix} 0 & 0 & f_{13} & f_{14} \\ 0 & 0 & f_{23} & f_{24} \\ f_{31} & f_{32} & f_{33} & f_{34} \\ f_{41} & f_{42} & f_{43} & f_{44} \end{bmatrix}. \quad (16)$$

The resulting linear push-broom epipolar equation reads as

$$(u_1'v_1', u_1', v_1', 1)\mathbf{F}(u_1v_1, u_1, v_1, 1)^T = 0. \quad (17)$$

We must point out that this $4 \times 4$ linear PB essential matrix result is not new; paper [6] already reported it though via a different approach. This however precisely confirms that our method provides a unified framework for handling different types of novel, higher-order epipolar geometries, including a PB camera.

**Difference with X-slit camera** An X-Slit camera collects rays that simultaneously pass through two oblique (neither parallel nor coplanar) slits in 3D space [28]. The linear PB camera give rise to two oblique slits setting, where one slit is the line of center of projection and the other slit corresponds to the viewing direction. However, the slit corresponds to the moving camera projection center is unknown and to estimate. Although the linear PB essential matrix shares the same size as the X-Slit camera essential matrix [25], the detailed structure is different.

### 4.4. A $6 \times 6$ essential matrix for uniform PB cameras

Similarly, for the uniform PB camera where the view plane of the camera is undergoing a uniform rotation besides its linear sweeping, the epipolar geometry can be represented as:

$$\left[0, v_i^{'}, 1\right] \left([\mathbf{t}]_\times \mathbf{R}_{u_i u_i^{'}} - u_i \mathbf{R}_{u_i u_i^{'}}[\mathbf{v}_1]_\times + u_i^{'}[\mathbf{v}_2]_\times \mathbf{R}_{u_i u_i^{'}}\right) \begin{bmatrix} 0 \\ v_i \\ 1 \end{bmatrix} = 0. \quad (18)$$

we can easily derive a $6 \times 6$ uniform PB essential matrix as:

$$\begin{bmatrix} u_i^{'2}v_i^{'} \\ u_i^{'2} \\ u_i^{'}v_i^{'} \\ u_i^{'} \\ v_i^{'} \\ 1 \end{bmatrix}^T \begin{bmatrix} 0 & 0 & f_{13} & f_{14} & f_{15} & f_{16} \\ 0 & 0 & f_{23} & f_{24} & f_{25} & f_{26} \\ f_{31} & f_{32} & f_{33} & f_{34} & f_{35} & f_{36} \\ f_{41} & f_{42} & f_{43} & f_{44} & f_{45} & f_{46} \\ f_{51} & f_{52} & f_{53} & f_{54} & f_{55} & f_{56} \\ f_{61} & f_{62} & f_{63} & f_{64} & f_{65} & f_{66} \end{bmatrix} \begin{bmatrix} u_i^2 v_i \\ u_i^2 \\ u_i v_i \\ u_i \\ v_i \\ 1 \end{bmatrix} = 0. \quad (19)$$

There are 32 variables in this PB essential matrix ($6 \times 6$ minus the top-left $2 \times 2$ corner), suggesting that a 31-point linear algorithm can be used to estimate $\mathbf{F}$. Note also that the resulting (generalized) epipolar curves are cubic.

**RS camera VS PB camera:** Both RS camera and PB camera have a scanline dependent pose, *i.e.*, temporal-dynamic center of projection. For PB cameras, the scanline direction is fixed relative to the local coordinate while the scanline direction changes with respect to the local coordinate for RS cameras. This creates the main difference between PB cameras and RS cameras and the extras freedom explains the increased order of polynomials in expressing the generalized epipolar geometry (4 VS 6 and 5 VS 7).

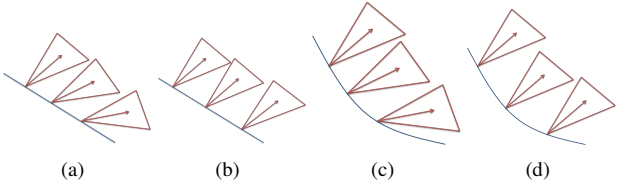

Figure 3. Camera models discussed in this paper. (a) Linear rolling shutter camera; (b) Linear push broom camera; (c) Uniform rolling shutter camera; (d) Uniform push broom camera.

### 4.5. Difference with "Multi-View Geometry for General Camera Models" [25]

In [25], Sturm derived the essential matrices for different general camera models by using the Plücker coordinates. For axial cameras with finite axis and with infinite axis, the essential matrices are of size $5 \times 5$, which is a reduced form the general $6 \times 6$ form. For X-slit cameras with two finite axes and with one finite and one infinite axis, the essential matrices are of size $4 \times 4$, which is another reduced form of the general $6 \times 6$ form for non-central cameras. Note that for these general cameras, the intrinsic configurations are fixed and known, *i.e.*, the axes for the axial camera and the X-slit camera. Therefore these essential matrices only encode the relative pose, *i.e.*, $\mathbf{R}, \mathbf{t}$.

By contrast, the rolling shutter cameras and push broom cameras are temporally dynamic, *i.e.*, the instantaneous motions are unknown and needed to estimate. Therefore, for these temporally dynamic cameras, we have to estimate not only the global motion ($\mathbf{R}, \mathbf{t}$) but also the the instantaneous motion ($\mathbf{w}, \mathbf{d}$).

## 5. Linear N-point algorithms for RS cameras

**Summary of the Above Results.** The above results are summarized in Table-1. We also include the number of points needed to solve *linearly* for the respective generalized essential matrices. Next, let us use as an example the linear RS camera to derive a **linear 20-point algorithm** for solving the uniform RS essential matrix. The linear solutions for other types of cameras in the table can be similarly derived, hence are omitted here. Interested readers will find more information in our supplementary material.

A short digest of the results is provided in Table-2.

Table 2. Number of points required for solving a generalized essential matrix by a linear method (first row,*i.e.* $\#(point)$), and by minimal solvers (second row, *i.e.* minimum-DOF).

| Camera | pinhole | lin-PB | lin-RS | uni-PB | uni-RS |
|---|---|---|---|---|---|
| #(points) | 8 | 11 | 20 | 31 | 44 |
| min.DOF | 5 | 11 | 11 | 17 | 17 |

### 5.1. A linear 20-point algorithm for RS cameras

For solving the linear RS relative pose problem, we first solve for the $5 \times 5$ RS essential matrix $\mathbf{F} \in \mathbb{R}^{5 \times 5}$. Then from its 21 non-zero elements, we recover the three atomic essential matrices $\mathbf{E}_0, \mathbf{E}_1$ and $\mathbf{E}_2$. Finally, the relative pose ($\mathbf{R}, \mathbf{t}$) and velocities $\mathbf{d}_1, \mathbf{d}_2$ can be simply extracted by decomposing $\mathbf{E}_0, \mathbf{E}_1$ and $\mathbf{E}_2$. The twisted pair ambiguity can be resolved by a standard method [8].

### 5.1.1 Solving the $5 \times 5$ linear RS essential matrix

The linear RS essential matrix $\mathbf{F}$ contains only 21 non-trivial homogeneous variables, hence its degree of freedom is 20. Collecting 20 correspondences, one can solve for the $5 \times 5$ matrix $\mathbf{F}$ linearly by SVD.

### 5.1.2 Recovering atomic essential matrices

Once the $5 \times 5$ matrix $\mathbf{F}$ is found, our next goal is to recover the individual atomic essential matrices $\mathbf{E}_0, \mathbf{E}_1$ and $\mathbf{E}_2$. Eq.-(11) provides 21 linear equations on the three essential matrices. As the three essential matrices consist of 27 elements, we need six extra constraints to solve for $\mathbf{E}_0, \mathbf{E}_1$ and $\mathbf{E}_2$. To this end, we resort to the inherent constraints on standard $3 \times 3$ essential matrices, *e.g.* $\det(\mathbf{E}) = 0$ and $2\mathbf{E}\mathbf{E}^T\mathbf{E} - \mathbf{Tr}(\mathbf{E}\mathbf{E}^T)\mathbf{E} = 0$, since $\mathbf{E}_0, \mathbf{E}_1$ and $\mathbf{E}_2$ are standard $3 \times 3$ essential matrices.

**A quadratic solution.** Examining the relationship between the linear RS essential matrix and the atomic essential matrices, we find that the right bottom $2 \times 2$ corner of $\mathbf{E}_0$ matrix can be directly read out; the first and second columns of $\mathbf{E}_1$ can also be read out; the first and second rows of $\mathbf{E}_2$ are also available from the RS essential matrix $\mathbf{F}$.

Taking $\mathbf{E}_1$ as an example, we now illustrate how to complete its missing column from two recovered columns. Once we have solved $\mathbf{F}$, we can directly read out the first two columns of $\mathbf{E}_1$, *i.e.* $\mathbf{E}_1 = \begin{bmatrix} E_1^{11} & E_1^{12} & * \\ E_1^{21} & E_1^{22} & * \\ E_1^{31} & E_1^{32} & * \end{bmatrix}$.
In order to recover the last missing column, we use both of its rank-2 constraint and cubic constraints. First, by using the rank-2 constraint we express the third column as a linear combination of the first two columns, *i.e.* $\begin{bmatrix} E_{13} \\ E_{23} \\ E_{33} \end{bmatrix} = \lambda_1 \begin{bmatrix} E_{11} \\ E_{21} \\ E_{31} \end{bmatrix} + \lambda_2 \begin{bmatrix} E_{12} \\ E_{22} \\ E_{32} \end{bmatrix}$. The remaining nonlinear constraints on a $3 \times 3$ essential matrix provide 9 equations over $\lambda_1$ and $\lambda_2$, among which we need to choose two in order to solve for $\lambda_1$ and $\lambda_2$. For simplicity, we only choose two quadratic ones, namely
$\begin{cases} a_{11}\lambda_1^2 + a_{12}\lambda_1\lambda_2 + a_{13}\lambda_2^2 + a_{14} = 0 \\ a_{21}\lambda_1^2 + a_{22}\lambda_1\lambda_2 + a_{23}\lambda_2^2 + a_{24} = 0 \end{cases}$. These quadratic equations can be solved efficiently by using any off-the-shelf solver. Following a similar procedure, we can also solve for $\mathbf{E}_2$, and subsequently solve for $\mathbf{E}_0$.

### 5.1.3 Recovering relative pose and velocities

Given the three essential matrices $\mathbf{E}_0 = [\mathbf{t}]_\times \mathbf{R}$, $\mathbf{E}_1 = [\mathbf{d}_1]_\times \mathbf{R}$, and $\mathbf{E}_2 = [\mathbf{d}_2]_\times \mathbf{R}$, we decompose them into relative transformations $(\mathbf{R}, \mathbf{t})$ and velocities $\mathbf{d}_1, \mathbf{d}_2$ [8]. During the above computation, the common rotation constraint has not been enforced explicitly, we could further introduce a non-linear optimization to enforce all the constraints.

## 5.2. Other linear solvers

**Linear 44-point algorithm for uniform RS camera.** For the uniform rolling shutter relative pose problem, we first solve for the uniform rolling shutter essential matrix $\mathbf{F} \in \mathbb{R}^{7 \times 7}$. Then from the 45 elements in $\mathbb{M}$, recover the eight matrices $\mathbf{E}_i, i = 0, \cdots, 7$. Finally, the relative pose $(\mathbf{R}, \mathbf{t})$ and velocities $\mathbf{v}_1, \mathbf{v}_2$ are extracted from the eight matrices. Due to its special structure, the uniform RS essential matrix $\mathbf{M}$ consists of 45 homogeneous variables, *i.e.*, 44 DoF. According to the uniform RS essential matrix Eq.-(15), by collecting 44 correspondences, we can solve for the uniform RS essential matrix $\mathbf{M}$ linearly through the singular value decomposition (SVD).

## 5.3. Normalization

In solving the linear RS essential matrix $\mathbf{F}$, it is important to implement a proper normalization: 1) Normalizing the image coordinates data $(u_i, v_i)$ and $(u'_i, v'_i)$ in the way as described in [7]. 2) Under the linear rolling shutter relative pose formulation, the inputs are monomials $(u_i^2, u_i v_i, u_i, v_i, 1)$ and $(u_i'^2, u'_i v'_i, u'_i, v'_i, 1)$, a better normalization should be defined on $(u_i^2, u_i v_i, u_i, v_i, 1)$ and $(u_i'^2, u'_i v'_i, u'_i, v'_i, 1)$ rather than $(u_i, v_i)$ and $(u'_i, v'_i)$. Therefore, we propose to normalize $(u_i^2, u_i v_i, u_i, v_i, 1)$ and $(u_i'^2, u'_i v'_i, u'_i, v'_i, 1)$ in the way as in [7].

## 6. Nonlinear Solvers w/ Sampson Error

Based on the above generalized essential matrices, we can now also devise nonlinear solvers. Instead of minimizing an algebraic error, we minimize the geometrically more meaningful (generalized) Sampson error metric. For example, in the case of a uniform RS camera, the Sampson error is the first-order approximation of the distance between a (generalized) feature vector $\mathbf{x}_i = [u_i^3, u_i^2 v_i, u_i^2, u_i v_i, u_i, v_i, 1]^T$ and its corresponding RS epipolar curve, *i.e.*,

$$e_{\text{Sampson}} = \sum_{i=1}^n \frac{(\mathbf{x}_i'^T \mathbf{F} \mathbf{x}_i)^2}{\sum_{j=1}^7 ((\mathbf{F}\mathbf{x}_i)_j^2 + (\mathbf{F}^T \mathbf{x}_i')_j^2)}. \quad (20)$$

We envisage three scenarios where such nonlinear solvers can prove useful:

a) Use it as a 'gold-standard' nonlinear refinement procedure.

b) Use it as a general solver which directly searches the variables.

c) Use it as a minimal-case solver together with RANSAC.

The last case is particularly relevant as RANSAC favors smaller sample sizes. For example, for the case of uniform RS camera our linear algorithm asks for 44 points; in contrast, a minimal-case solver only requires 17 points, as there are in total 18 degrees of freedom in $[\mathbf{R}, \mathbf{t}, \mathbf{w}_1, \mathbf{w}_2, \mathbf{d}_1, \mathbf{d}_2]$.

To solve the above Sampson error minimization problem, we parametrize the rotation with its angle-axis representation, then we use the standard unconstrained optimization solver 'fminunc' in Matlab.

## 7. Experimental evaluation

We evaluated the linear and uniform RS relative pose methods on both synthetic and real image datasets. When ground-truth data is available, error metrics for rotation and translation estimates are defined as $e_R = \mathrm{acos}((\mathrm{trace}(\widehat{\mathbf{R}}\mathbf{R}_{GT}^T) - 1)/2)$, and $e_T = \mathrm{acos}(\widehat{\mathbf{t}}^T \mathbf{t}_{GT}/(\|\widehat{\mathbf{t}}\|\|\mathbf{t}_{GT}\|))$.

### 7.1. Simulation Experiments

Generating geometrically consistent simulation measurements for a dynamic RS camera is a challenge in itself. First, a relative pose $(\mathbf{R}, \mathbf{t})$ is randomly defined between the image pair. The focal length is set to 640 while the image resolution is defined to be $640 \times 480$. Second, given translation velocities $\mathbf{d}_1$ and $\mathbf{d}_2$, and angular velocities $\mathbf{w}_1, \mathbf{w}_2$, the camera pose for each row can be determined. The correspondences are then simulated such that they are not too far from what a real world image feature tracker would return. Each generation is finalized by a cheirality check to guarantee the corresponding 3D point lies in front of both cameras. All experiments are repeated 200 times to obtain statistically relevant conclusions.

**Evaluation of the linear methods.** Here we first test our 20-point algorithm for linear RS relative pose. We use the angle between vectorized ground truth and estimated essential matrices as a performance indicator. Fig. 4 illustrates the essential matrix estimation error with respect to increasing noise. The figure is using a double logarithmic scale. For the 44-point algorithm, a similar curve could also be obtained. We observe that the linear methods are very sensitive to noise. To deal with real world noise, in the following experiments, we used the nonlinear optimization method.

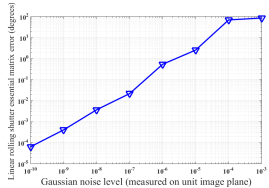

Figure 4. Evaluation on increasing Gaussian noise for linear 20-point algorithm. Noise is added to the normalized coordinates.

**Accuracy versus noise level.** To evaluate the performance in the presence of noise, we added random Gaussian noise to the correspondences. As we worked mainly on the normalized image coordinates, noise was added immediately on the normalized image plane (*i.e.*, unit image plane). Statistical results are illustrated in Fig. 5, demonstrating that our linear RS camera model always achieves better performance than the global shutter camera model, while both rotation and translation errors increase with increasing noise level.

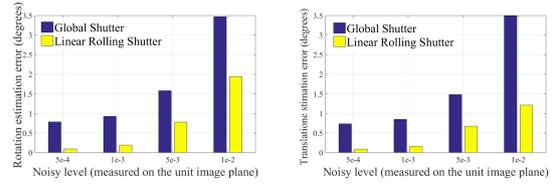

(a) Rotation estimation error  (b) Translation estimation error

Figure 5. Performance evaluation with increasing Gaussian noise.

**Accuracy versus focal-length.** The observability of the RS effect depends on several factors, namely, focal length, depth of the 3D points and the ratio between linear and angular velocities. Here we investigate the performance of relative pose estimation with respect to the focal length. For a constant Gaussian noise level of $2 \times 10^{-3}$, we decrease the camera focal length from 640 to 80. Experimental results of rotation and translation estimation are illustrated in Fig.6. With a decreasing focal length, the RS effect becomes increasingly well observable, leading to a decrease of the motion estimation error. However, the pose estimation error does not necessarily decrease monotonically.

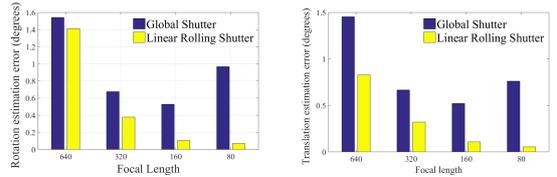

(a) Rotation estimation error  (b) Translation estimation error

Figure 6. Evaluation on decreasing focal length with noise of $2 \times 10^{-3}$ standard deviation on the unit image plane.

**Accuracy versus RS velocity.** Finally, we analyzed the effect of varying dynamics on the RS effect and the accuracy of the RS relative pose algorithm. We decreased the scale of the translation velocity from $10^{-2}$ to $10^{-4}$. The results are illustrated in Fig. 7. With an increasing velocity, our linear RS model achieves an obvious improvement in pose estimation, which suggests that the RS effect is more observable under large linear and angular motion.

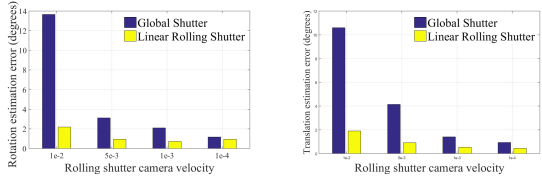

(a) Rotation estimation error   (b) Translation estimation error

Figure 7. Evaluation over decreasing translation velocity with noise $5 \times 10^{-3}$ standard deviation on the unit image plane.

### 7.2. Tests on synthetic RS images

To evaluate the performance of our RS relative pose solvers, we further used the simulated RS image datasets from [4]. This dataset includes six image sequences generated by 'Autodesk's Maya', where each sequence consists of 12 RS distorted frames. Ground truth camera poses were provided for each row of the image frame. These experiments allow full control over the rolling shutter effect, while at the same time representing a realistic scenario that allows for the application of a real feature tracker.

As pure rotation is always a degenerate case for epipolar geometry, we used only the last sequence "house_trans_rot1_B40" in our experiment, where the camera experiences both translational and angular displacements. To establish correspondences between the image frames, we used the standard KLT tracker (A sample result is shown in Fig. 8). Both global shutter camera model and uniform RS model were used to estimate the camera motion. In Fig. 9, we compare the accuracy of the resulting rotation estimation for the global shutter model and our uniform rolling shutter solution. Our method achieves a significant improvement on most of the image frames.

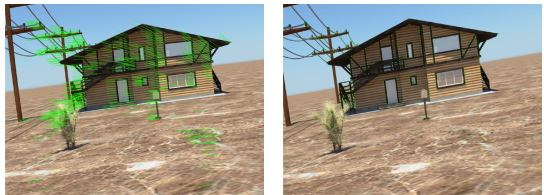

Figure 8. Synthetic image experiments on the sequence "house_trans_rot1_B40". (a) KLT tracking results, (b) Tracked features in the 2nd image. (**Best viewed in color.**)

### 7.3. Test on real RS images

We tested our algorithm on pairs of images taken from a publicly available RS images dataset (http://www.cvl.isy.liu.se/research/datasets/rsba-dataset/). The pairs are chosen such that the median frame-to-frame disparity of the extracted feature correspondences remains below 100 pixels. The images have a resolution of 1280×720, and are captured by an iPhone 4 camera. The focal length of the camera is 1485.2, and the principal point is simply defined as the center of the image. We apply a Harris corner extractor and 31×31 image patches to extract the interest points, and match them using a simple brute-force approach. We apply Ransac to the resulting correspondences, and refine the final model over all inliers. In each iteration, we first apply a global shutter relative pose solver to identify all inliers and initialize the relative pose, and then use Sampson error minimization in order to optimize the result. We use standard Sampson error minimization (*i.e.* based on a global shutter model) as a baseline implementation, and our adapted Sampson error for RS cameras as the improved alternative.

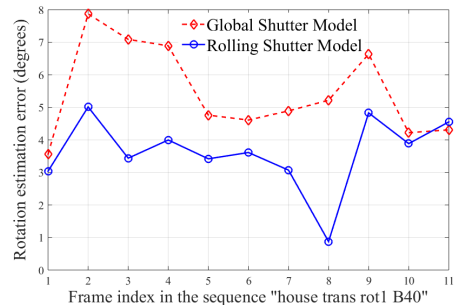

Figure 9. Rotation estimation performance comparison between global shutter model and our uniform rolling shutter solver.

An example result with 2287 input correspondences is shown in Figure 10. As can be clearly observed, the RS model allows for a more complete description of the geometry, and leads to a significant reduction in the (approximate) reprojection error after the final optimization step. Moreover, it is interesting to see that the global shutter model achieves a relatively small error for a sub-part of the image only, while the RS model is able to explain the distortion in other regions and achieves a small error in almost the entire image. A similar difference in performance can be observed for any pair of images with sufficient dynamics, thus underlining the importance of taking the RS effect into account.

## 8. Conclusion

We have derived novel generalized essential matrices of size $4 \times 4$, $5 \times 5$, $6 \times 6$, and $7 \times 7$ for linear PB, linear RS, uniform PB, and uniform RS cameras, respectively. We also developed effective linear N-point algorithms and non-linear Sampson error minimizers for solving these generalized essential matrices. The entire work represents a unified and elegant framework for solving the Relative Pose problem with new types of cameras, including the practically relevant and previously unsolved case of a RS camera. It is our hope that the presented theoretical contribution to the field of epipolar geometry will serve as a solid foundation for further extensions to novel and practically relevant types of cameras. This, for instance, includes light-field cameras [12], general linear cameras [27], and generalized camera models [19, 14, 25]. The theory promises a more general applicability to spatio-temporally scanning sensors, such as satellite imagery and sweeping Laser scanners.

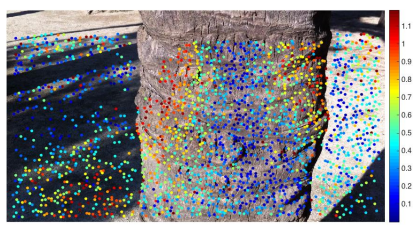

(a) Global shutter model

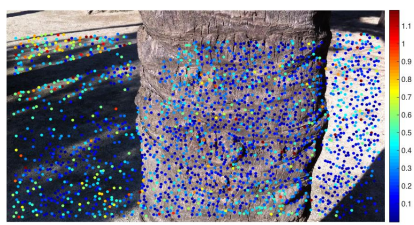

(b) Rolling shutter model

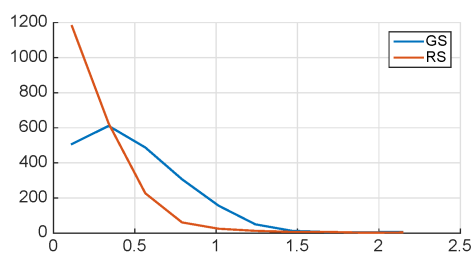

(c) Histogram of Sampson errors

Figure 10. Comparisons of the Sampson errors for a pair of images taken from a RS video dataset. (a) shows the final result of Sampson error minimization based on a global shutter model. The error distribution has a structure in the image plane, indicating regions for which the RS distortion is not properly taken into account. (b) shows how the inclusion of a RS model and the extended Sampson distance take those distortions into account, and produce a reprojection error that distributes much more uniformly across the entire image plane. (c) illustrates a histogram of reprojection errors for both cases, thus demonstrating a general reduction of the error through the used of the proposed rolling shutter essential matrix.

## Acknowledgments

Y. Dai is funded by ARC Grants (DE140100180, LP100100588) and National Natural Science Foundation of China (61420106007). H. Li's research is funded in part by ARC grants (DP120103896, LP100100588, CE140100016) and NICTA (Data61). L. Kneip is funded by ARC grants DE150101365 and CE140100016.

## References

[1] O. Ait-Aider, N. Andreff, J. Lavest, and P. Martinet. Simultaneous object pose and velocity computation using a single view from a rolling shutter camera. In *Proc. Eur. Conf. Comp. Vis.*, pages 56–68. 2006. 2

[2] O. Ait-Aider and F. Berry. Structure and kinematics triangulation with a rolling shutter stereo rig. In *Proc. IEEE Int. Conf. Comp. Vis.*, pages 1835–1840, Sept 2009. 3

[3] C. Albl, Z. Kukelova, and T. Pajdla. R6p - rolling shutter absolute camera pose. In *Proc. IEEE Conf. Comp. Vis. Patt. Recogn.*, June 2015. 1, 2

[4] P.-E. Forssen and E. Ringaby. Rectifying rolling shutter video from hand-held devices. In *Proc. IEEE Conf. Comp. Vis. Patt. Recogn.*, pages 507–514, June 2010. 9

[5] M. Grundmann, V. Kwatra, D. Castro, and I. Essa. Calibration-free rolling shutter removal. In *ICCP*, pages 1–8, April 2012. 2

[6] R. Gupta and R. Hartley. Linear pushbroom cameras. *IEEE Trans. Pattern Anal. Mach. Intell.*, 19(9):963–975, Sep 1997. 2, 5

[7] R. Hartley. In defense of the eight-point algorithm. *IEEE Trans. Pattern Anal. Mach. Intell.*, 19(6):580–593, Jun 1997. 7

[8] R. I. Hartley and A. Zisserman. *Multiple View Geometry in Computer Vision*. Cambridge University Press, ISBN: 0521540518, second edition, 2004. 6, 7

[9] J. Hedborg, P.-E. Forssen, M. Felsberg, and E. Ringaby. Rolling shutter bundle adjustment. In *Proc. IEEE Conf. Comp. Vis. Patt. Recogn.*, pages 1434–1441, June 2012. 1, 2

[10] J. Hedborg, E. Ringaby, P.-E. Forssen, and M. Felsberg. Structure and motion estimation from rolling shutter video. In *International Conference on Computer Vision Workshops*, pages 17–23, Nov 2011. 2

[11] S. Im, H. Ha, G. Choe, H.-G. Jeon, K. Joo, and I. S. Kweon. High quality structure from small motion for rolling shutter cameras. In *Proc. IEEE Int. Conf. Comp. Vis.*, Santiago, Chile, 2015. 1, 2

[12] O. Johannsen, A. Sulc, and B. Goldluecke. On linear structure from motion for light field cameras. In *Proc. IEEE Int. Conf. Comp. Vis.*, 2015. 10

[13] C. Kerl, J. Stueckler, and D. Cremers. Dense continuous-time tracking and mapping with rolling shutter RGB-D cameras. In *Proc. IEEE Int. Conf. Comp. Vis.*, Santiago, Chile, 2015. 1, 3

[14] H. Li, R. Hartley, and J.-H. Kim. A linear approach to motion estimation using generalized camera models. In *Proc. IEEE Conf. Comp. Vis. Patt. Recogn.*, pages 1–8, June 2008. 10

[15] L. Magerand, A. Bartoli, O. Ait-Aider, and D. Pizarro. Global optimization of object pose and motion from a single rolling shutter image with automatic 2d-3d matching. In *Proc. Eur. Conf. Comp. Vis.*, pages 456–469, 2012. 1, 2

[16] M. Meingast, C. Geyer, and S. Sastry. Geometric models of rolling-shutter cameras. In *OMNIVIS*, 2005. 2, 5

[17] L. Oth, P. Furgale, L. Kneip, and R. Siegwart. Rolling shutter camera calibration. In *Proc. IEEE Conf. Comp. Vis. Patt. Recogn.*, pages 1360–1367, June 2013. 3

[18] A. Patron-Perez, S. Lovegrove, and G. Sibley. A spline-based trajectory representation for sensor fusion and rolling shutter cameras. *Int. J. Comput. Vision*, 113(3):208–219, July 2015. 3

[19] R. Pless. Using many cameras as one. In *Proc. IEEE Conf. Comp. Vis. Patt. Recogn.*, pages 587–593, June 2003. 10

[20] J. Ponce. What is a camera? In *Proc. IEEE Conf. Comp. Vis. Patt. Recogn.*, pages 1526–1533, June 2009. 3

[21] O. Saurer, K. Koser, J.-Y. Bouguet, and M. Pollefeys. Rolling shutter stereo. In *Proc. IEEE Int. Conf. Comp. Vis.*, pages 465–472, Dec 2013. 1, 3

[22] O. Saurer, M. Pollefeys, and G. H. Lee. A minimal solution to the rolling shutter pose estimation problem. In *IEEE/RSJ International Conference on Intelligent Robots and Systems*, 2015. 1, 2

[23] Y. Sheikh, A. Gritai, and M. Shah. On the spacetime geometry of galilean cameras. In *Proc. IEEE Conf. Comp. Vis. Patt. Recogn.*, pages 1–8, June 2007. 5

[24] G. S. Steven Lovegrove, Alonso Patron-Perez. Spline fusion: A continuous-time representation for visual-inertial fusion with application to rolling shutter cameras. In *Proc. Brit. Mach. Vis. Conf.*, 2013. 3

[25] P. Sturm. Multi-view geometry for general camera models. In *Proc. IEEE Conf. Comp. Vis. Patt. Recogn.*, pages 206–212, June 2005. 5, 6, 10

[26] F. Vasconcelos and J. Barreto. Towards a minimal solution for the relative pose between axial cameras. In *Proc. Brit. Mach. Vis. Conf.*, 2013. 5

[27] J. Yu and L. McMillan. General linear cameras. In T. Pajdla and J. Matas, editors, *Proc. Eur. Conf. Comp. Vis.*, volume 3022 of *Lecture Notes in Computer Science*, pages 14–27. Springer Berlin Heidelberg, 2004. 2, 10

[28] A. Zomet, D. Feldman, S. Peleg, and D. Weinshall. Mosaicing new views: The crossed-slits projection. *IEEE Trans. Pattern Anal. Mach. Intell.*, 25(6):741–754, June 2003. 5